\documentclass[conference]{IEEEtran}
\IEEEoverridecommandlockouts
\usepackage{cite}
\usepackage{amsmath,amssymb,amsfonts}
\usepackage{algorithm}
\usepackage{algpseudocode}
\usepackage{graphicx}
\usepackage{textcomp}
\usepackage{xcolor}
\usepackage{comment}
\usepackage{tabularx,booktabs}
\usepackage{hyperref}
\def\BibTeX{{\rm B\kern-.05em{\sc i\kern-.025em b}\kern-.08em
    T\kern-.1667em\lower.7ex\hbox{E}\kern-.125emX}}

\usepackage{authblk}
\newcolumntype{C}{>{\centering\arraybackslash}X} 
\usepackage[margin=51.5pt]{geometry}

\author[1,2]{Thomas Manteaux}
\author[2]{David Rodríguez-Martínez}
\author[1]{Raj Thilak Rajan}  
\affil[1]{Signal Processing Systems, Faculty of EEMCS, Delft University of Technology, The Netherlands}
\affil[2]{Advanced Quantum Architecture Lab (AQUA), Ecole Polytechnique Fédérale de Lausanne (EPFL), Switzerland}
\date{}                     
\makeatletter
\def\@maketitle{%
  \newpage
  \null
  \vskip 72pt
  \begin{center}%
  \let \footnote \thanks
    {\LARGE \@title \par}%
    \vskip 1.5em%
    {\large
      \lineskip .5em%
      \begin{tabular}[t]{c}%
        \@author
      \end{tabular}\par}%
    \vskip 1em%
    {\large \@date}%
  \end{center}%
  \par
  \vskip 0.5em}
\makeatother

\begin{document}

\title{RAPF: Efficient path planning for lunar microrovers\\
}

\maketitle

\begin{abstract}
Efficient path planning is key for safe autonomous navigation over complex and unknown terrains. Lunar Zebro (LZ), a project of the Delft University of Technology, aims to deploy a compact rover, no larger than an A4 sheet of paper and weighing not more than $3$ kilograms. In this work, we introduce a Robust Artificial Potential Field (RAPF) algorithm, a new path-planning algorithm for reliable local navigation solution for lunar microrovers. RAPF leverages and improves state of the art Artificial Potential Field (APF)-based methods by incorporating the position of the robot in the generation of bacteria points and considering local minima as regions to avoid. We perform both simulations and on field experiments to validate the performance of RAPF, which outperforms state-of-the-art APF-based algorithms by over 15\% in reachability within a similar or shorter planning time. The improvements resulted in a 200\% higher success rate and 50\% lower computing time compared to the conventional APF algorithm. Near-optimal paths are computed in real-time with limited available processing power. The bacterial approach of the RAPF algorithm proves faster to execute and smaller to store than path planning algorithms used in existing planetary rovers, showcasing its potential for reliable lunar exploration with computationally constrained and energy constrained robotic systems. \\

\end{abstract}


\section{Introduction}


Despite significant advancements in space technology and robotics, navigating efficiently across unknown and unstructured environments remains an ongoing pursuit. This paper explores the importance of efficient path planning for small lunar rovers ($<$100\;kg) and microrovers ($<$10\;kg), which have gained significance with the advent of cheaper and commercial exploration missions over the last decade. 
Several state-of-the-art lunar rover path planning solutions have been developed, drawing from diverse fields such as robotics, artificial intelligence, and space exploration. For instance, Yu et al. \cite{Yu2020} propose a heuristic-based approach that considers terrain complexity and energy efficiency for global path planning on the lunar surface. Another notable work by Peng et al. \cite{Peng2023} utilizes deep neural networks to to perform data-driven path planning on the basis of heuristic search. These advancements showcase the ongoing efforts to enhance the autonomy and efficiency of lunar exploration missions through innovative path planning techniques. Although recent trends, such as deep learning, show promise, they remain computationally intensive, as compared to more resource efficient algorithms e.g., Rule-based approaches, including Dijkstra-based \cite{Dijkstra1959}, Artificial Potential Field (APF)-based \cite{Khatib1985}, and sampling-based \cite{LaValle1998} algorithms. Small lunar rovers, operating in complex environments with communication delays, restricted computational, memory and power capabilities, and limited sensor suites demand robust, real-time path planning algorithms. These algorithms must cope with the irregularities of the lunar surface---characterized by uneven and unconsolidated terrains filled with craters and rocks of different sizes---while addressing measurement noise from sensors on-board the rover. Due to their small size and reduced mobility, microrovers face challenges in negotiating rough or uneven terrain. 
\begin{figure}[t]
    \centering
    \includegraphics[width=1\linewidth]{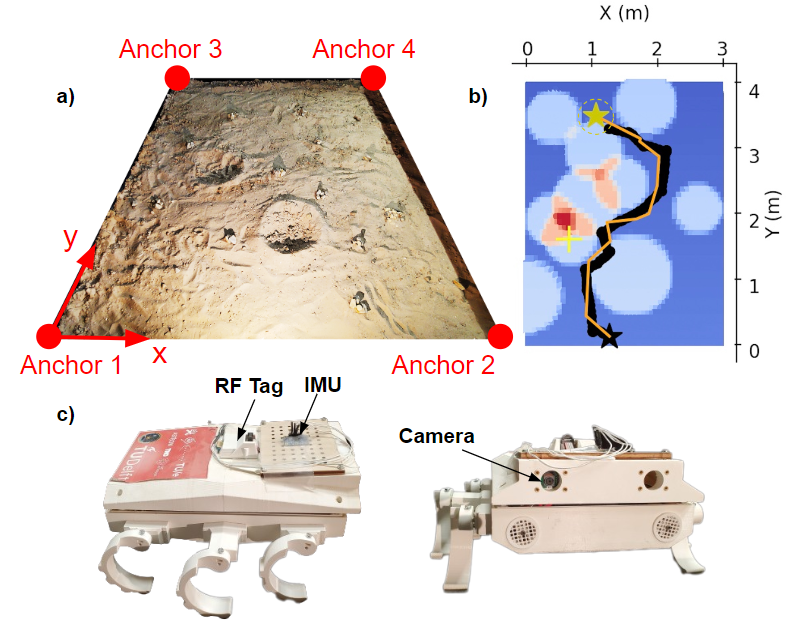}
    \caption{a) Field test layout. b) Potential map computed by the RAPF algorithm. Dark blue represents low-potential regions while red represents high-potential regions. The orange line shows the computed path, black line defines the ground truth, and the yellow cross shows where a local minimum was found. c) Lunar Zebro prototype used for testing}
    \label{fig:intro}
\end{figure}
Small size constrains sensor range, limiting the perception of surrounding environments for micro-rovers preventing comprehensive perception and posing challenges for effective path planning in navigation. The power constraint of lunar rovers demands low-complexity and efficient path planning, striking a balance between computational cost and performance. 
Path planning emerges, therefore, as a critical aspect impacting mission efficiency and success. 

\subsection{Contributions} To robustly and efficiently compute the path to be followed by a lunar microrover, we propose a new lightweight solution, RAPF, based on adapting and expanding an existing APF-based algorithm. Our approach is adapted to run onboard the LZ walking rover on a Raspberry Pi 4B. We demonstrate the performance of our algorithm both in simulations and during field tests across a lunar-like environment. Results show the greater accuracy and lower memory allocation requirements of our approach, displaying comparable performances even when compared to optimal terrestrial algorithms such as A*. 

\subsection{Related work} Different approaches exist for computing the optimal path to follow between points of interest.

\textit{Combinatorial methods}---which include the Dijkstra, A* \cite{Dijkstra1959}, and D* \cite{Hart1968} algorithms---explore discrete paths through a graph representation of the environment. Graph nodes represent specific configurations of the robot. Edges represent feasible transitions between configurations. Only transitions between neighboring graph nodes are allowed. Graph search algorithms explore the graph by visiting nodes and searching for the optimal path based on certain criteria such as distance or cost. Despite ensuring optimal paths and completeness, combinatorial methods demand significant memory and computational capabilities, struggling in environments where the map is frequently recomputed.

\textit{Sampling-based algorithms} divide a continuous space into sample points forming a roadmap or tree, with Rapidly Exploring Random Tree (RRT) \cite{LaValle1998} and Probabilistic Roadmaps (PRM) \cite{Kavraki1996} as prominent examples of this approach. Sampling involves a deterministic or stochastic selection of points within the configuration space, representing potential robot configurations. These sampled points stand as potential configurations that the robot can take. The objective is to comprehensively cover the configuration space to unveil feasible paths. The connection of these points forms a graph structure, balancing optimality and completeness. However, these algorithms can be sensitive to the initial configuration \cite{Karaman2011} and may necessitate time-intensive pre-processing.

Biologically inspired methods such as \textit{spiking neural models} \cite{Culloch1943} use artificial neural networks to mimic the brain's structure, enabling robots to adaptively plan paths based on sensory inputs. Swarm intelligence, inspired by collective behavior, uses algorithms like Particle Swarm Optimization \cite{Kennedy1965} and Ant Colony Optimization \cite{Dorigo2006}. Genetic algorithms \cite{Katoch2021}, derived from natural selection principles \cite{Holland1992}, iteratively refine potential solutions. As they replicate problem-solving strategies found in nature, biologically inspired methods offer a large adaptability making them well-suited in complex and dynamic environments. They provide robustness against unpredicted obstacles and changing surroundings. However, they require a large number of parameters to be fine-tuned and their implementation is usually complex. They therefore have a large computational cost. Biologically inspired methods do not always outperform traditional path planning methods -- combinatorial, sampling-based and APF-based methods -- in static environments and their complexity might hinder their implementation on embedded systems working in real-time.

Lastly, potential field techniques \cite{Khatib1985} guide robots using artificial forces. In the \textit{Artificial Potential Field} (APF) method, an attractive potential pulls the robot towards the target, while repulsive potentials keep it away from obstacles. The APF approach ensures collision-free paths in real-time with low computational complexity. However, they face local minima issues, potentially trapping the robot in sub-optimal positions. Local minimas occur due to various reasons such as symmetry between the different obstacles, closely spaced obstacles, or in highly cluttered environments. Tuning the balance between repulsive and attractive potentials is crucial for successful path planning and to avoid local minima. The next section presents the original APF algorithm and some of its more promising variations. These form the basis of the approach proposed herein.

\section{Artificial Potential Field (APF) algortihms}


The classical APF was originally proposed in \cite{Khatib1985}, in which a robot navigates in an artifical force field, with attractive forces being emitted by the target and repulsive forces by the obstacles. These forces are directly derived from quadratic potential functions, which are parameterized by the robots distance to a target or an obstacle. The robot, therefore, aims at minimizing its potential cost function at any given point along its' path. A local minima event occurs,  when no adjacent point exists in the neighborhood of the rover with a lower potential as compared to the current position of the robot. The attractive potential acting on a robot at the global position  $\textbf{x} = [x,y]^T$ in a 2-dimensional scenario, from a target located at  $\textbf{x}_t = [x_t,y_t]^T$ is defined as
\begin{equation} \label{eq:Ja}
    J_a(\textbf{x}) = \frac{1}{2}k_a d_t^2(\textbf{x}),
\end{equation}
where $k_a$ is the attractive gain, $d_t(\textbf{x}) = ||\textbf{x}-\textbf{x}_t||_2$ is the Euclidean distance between the robot and the target. The repulsive potential acting on a robot located at $\textbf{x}$ from the \textit{n}th detected obstacle at location $\textbf{x}_o^n = [x_o^n,y_o^n]^T$ is defined as
\begin{equation} \label{eq:Jo}
    J_o^n(\textbf{x}) =
    \begin{cases}
     \frac{1}{2}k_r(\frac{1}{d_n(\textbf{x})}-\frac{1}{\rho_0})^2 & \text{if $d_n(\textbf{x}) \leq \rho_0$}, \\
      0 & \text{otherwise},
    \end{cases}
\end{equation} where $k_r$ is the repulsive gain, $\rho_0$ the maximum distance within which the potential field of the obstacle exerts its influence, and $d_n(\textbf{x}) = ||\textbf{x}-\textbf{x}_o^n||_2 $ is the Euclidean distance between the robot and the \textit{n}th detected obstacle.
Let $N_{obs}$ be the number of obstacles detected by the robot at position \textbf{x}, then the total repulsive potential acting on the robot is
\begin{equation}
    J_r(\textbf{x}) = \sum_{n=1}^{N_{obs}} J_o^n(\textbf{x})
    \label{eq:Jr}
\end{equation}
and the total potential function is the sum of the attractive and the repulsive potentials as stated below
\begin{equation} \label{eq:J}
    J(\textbf{x}) = J_a(\textbf{x}) + J_r(\textbf{x}).
\end{equation}
The robot's direction of motion is determined by calculating the gradient relative to the potential of its current position.

\begin{table*}[htbp]
\caption{Overview of state-of-the-art APF-based algorithms}
\label{my-tab:apf_algos}
\begin{tabularx}{\textwidth}{@{} l *{4}{C}}
\toprule
    \textbf{Algorithm} & APF & RVF & CRBAPF* & RAPF\\
    \midrule
    \textbf{Key concept} & Attractive/Repulsive potential & Rotated vector field & Bacteria points + RWT & Bacteria points + artificial obstacles \\
    \textbf{Cost function type} & Quadratic & Quadratic & Gaussian & Gaussian \\
    \textbf{Handling Local minima} & - & Rotation of the field & Random walk technique & Artificial obstacles \\
    \textbf{Trajectory smoothness} & Abrupt changes & Smooth trajectories & Oscillations for low $N_B$ & Smooth trajectories\\
\bottomrule
\end{tabularx}
\end{table*}

\subsection{RVF} An alternative potential field method for escaping local minima is proposed in \cite{Nasuha2022} and \cite{Gelling2023}, called the Rotated Vector Field (RVF), which reorients the force field directly derived from the classical APF potential functions as described below. The resulting attractive force is defined as the gradient function of the attractive potential
\begin{equation}
    \textbf{F}_a(\textbf{x}) = - \nabla J_a(\textbf{x}),
\end{equation} where $J_a(\textbf{x})$ is defined in (\ref{eq:Ja}). Similar to the attractive force, the resulting repulsive force is directly derived from the potential function  as
\begin{equation}
    \textbf{F}_r^n(\textbf{x}) = - \nabla J_o^n(\textbf{x}),
\end{equation}
\noindent where $J_o^n(\textbf{x})$ is defined in (\ref{eq:Jo}). Similar to the total force field defined for APF in (\ref{eq:J}), the total force is the sum of the attractive and repulsive forces written as
\begin{equation}
    \textbf{F}(\textbf{x}) = - \nabla J(\textbf{x}) = - (\nabla J_a(\textbf{x}) + \nabla J_r(\textbf{x})).
\end{equation} The main benefit of this approach is that instead of being repelled when approaching an obstacle, the robot is forced to move around it. This is achieved by rotating the vector field around the obstacles in one direction or another. 
This creates a vortex field at the location of the \textit{n}th detected obstacle given by
\begin{equation} \label{eq:rvf_field}
   F_{n}(\textbf{x}) = \pm
    \begin{pmatrix}
        \frac{\partial J_{t,n}(\textbf{x})}{\partial y} \\
        \frac{-\partial J_{t,n}(\textbf{x})}{\partial x} \\
    \end{pmatrix}.
\end{equation}
The intensity of the vector field remains the same, only the direction changes. The vortex field is bounded at a distance $\rho_0$ from the obstacles, the maximum distance within which the potential field of the obstacle exerts its influence, such that the entire vector field does not take on this rotary shape.

\subsection{CRBAPF*} In \cite{Hossain2015} and \cite{Ahmad2020} a set of enhanced APF algorithms called Bacteria APFs (BAPFs) are proposed. These algorithms involve what is called \textit{bacteria points}, i.e., virtual points around the robot representing a potential solution to the optimization problem of the APF algorithm. A  new APF algorithm based on the BAPF concept was introduced in \cite{Rajan2023}, named Changing Radii Bacteria APF (CRBAPF*), which actively modifies the potential functions and enables a random walk technique (RWT) to escape local minima. The potential functions are computed for the bacteria points by taking into account their distances from the target and any obstacles detected in the environment. The bacteria with the lowest cost is chosen as the next position for the robot, and the latter moves towards that selected point. The cost functions from the target and the obstacles are expressed as Gaussian functions instead of quadratic functions in the classical APF method. Using Gaussian over quadratic cost functions presents several advantages such as a more gradual increase in repulsion as the robot approaches an obstacle and a smoother transition from attractive to repulsive forces. This solution results in smoother trajectories, a more natural behavior, and a better handling of narrow passages. The attractive potential acting on a robot located at $\textbf{x} = [x,y]^T$ from a target located at $\textbf{x}_t = [x_t,y_t]^T$ is defined as
\begin{equation}
    J_a(\textbf{x}) = -\alpha_ae^{-\mu_ad_t(\textbf{x})},
\end{equation}
with $\alpha_a$ and $\mu_a$ constant parameters established through empirical means and they can be seen as the height and the width of the potential source. $d_t(\textbf{x})$ is the squared Euclidean distance between the robot and the target. 
The repulsive potential acting on a robot located at \textbf{x} from the \textit{n}th detected obstacle at location $\textbf{x}_o^n = [x_o^n,y_o^n]^T$ is defined as
\begin{equation} \label{eq:cr_function}
    J_o^n(\textbf{x}) =
    \begin{cases}
      0 & \text{if $\sqrt{d_n(\textbf{x})} > \rho_u$}, \\
      \alpha_oe^{-\mu_o d_n(\textbf{x})} & \text{if $\rho_l <\sqrt{d_n(\textbf{x})} < \rho_u$}, \\
      \infty & \text{if $\sqrt{d_n(\textbf{x})} < \rho_l$},
    \end{cases}
\end{equation}
with $\rho_l$ the lower radius and $\rho_u$ the upper radius setting the boundaries of the potential source. $\alpha_o$ and $\mu_o$ are constants tuned by testing and dependent on the average density of obstacles, and $d_n(\textbf{x})$ is the squared Euclidean distance between the robot and the \textit{n}th detected obstacle. For a number $N_{obs}$ of obstacles detected by the robot at a position \textbf{x}, the total repulsive potential acting on the robot is \\
\begin{equation}
    J_r(\textbf{x}) = \sum_{n=1}^{N_{obs}} J_o^n(\textbf{x}),
\end{equation}
and the total potential function is the sum of the attractive and the repulsive potentials as stated below.
\begin{equation}
\label{eq:total_potential}
    J_r(\textbf{x}) = J_a(\textbf{x}) + J_r(\textbf{x}).
\end{equation}

\noindent \textit{Bacteria potential}

For a robot located at $\textbf{x} = [x,y]^T$, the position of the \textit{j}th the bacteria is $\textbf{x}_{b,j} = [x_{b,j},y_{b,j}]^T$, defined as 
 \begin{equation}
    \begin{pmatrix}
           x_{b,j} \\
           y_{b,j} \\
    \end{pmatrix}
    =
    \begin{cases}
      x + \rho_b cos(\frac{2\pi n_j}{N_B}), \\
      y + \rho_b sin(\frac{2\pi n_j}{N_B}),
    \end{cases}
\end{equation}
where $N_B$ is the number of bacteria, $n_j = 1,2,...,N_B$ and $\rho_b$ the radius around the robot where bacteria are generated. Similarly to the rover, the total potential function at the location of the \textit{j}th bacteria is expressed as
\begin{equation}
    J_{b,j}(\textbf{x}_{b,j}) = J_a(\textbf{x}_{b,j}) + J_r(\textbf{x}_{b,j}).
\end{equation}
As the robot is looking for the lowest potential towards the target, the next selected bacteria point \textit{j} shall meet the criterion
\begin{equation} \label{eq:b_criteria}
    J_{b,j}(\textbf{x}_{b,j}) - J_r(\textbf{x}) < 0.
\end{equation}
If several bacteria points meet (\ref{eq:b_criteria}), the one with the lowest Euclidean distance toward the target is selected. \\

\noindent \textit{Random walk technique} \\
To prevent the robot from getting stuck in a local minimum, a random walk technique is employed. The robot randomly selects a bacteria point around it, moves to it, and repeats the process for a pre-defined amount of time. The random selection of the bacteria point follows a uniform distribution. The only condition to be met follows from (\ref{eq:cr_function}).

\subsection{Robust and Lightweight APF}

Our proposed approach in this work, the Robust and Lightweight APF (RAPF) method, differs from the CRBAPF* algorithm in two different ways. First, RAPF uses a different local minima escape approach instead of the random walk technique. In RAPF, when local minimas are encountered, then they are labeled as artificial obstacles. These artificial obstacles generate a repulsive potential (Eq. \ref{eq:cr_function}) that influences the total potential function (Eq. \ref{eq:total_potential}). The existing planned path is cleared and a new path is computed from the current position, which includes the local minima as an obstacle. Another improvement introduced into the RAPF algorithm consists of taking into account the position of the robot with respect to the target during the generation of the bacteria points. The $N_B$ bacteria points are generated such that there is always a bacteria point along the robot-target vector. This removes the oscillation effect from the CRBAPF* algorithm when a small number of bacteria points are used. This feature enables the RAPF algorithm to use a reduced number of bacteria points ($N_B$ = 8), resulting in a more lightweight method. Algorithm \ref{alg:RAPF} describes the working principle of the RAPF algorithm. 

At the onset crucial parameters are set-up like observed obstacles, the number of bacterial agents, movement step size, and maximum duration. It initializes key variables such as the current position, a timer for elapsed time tracking, and a boolean flag for path determination. Within each iteration, the algorithm assesses the robot's position by calculating the cost function (J) considering obstacle influence. It computes the cost function for each bacteria point ($Idx_B$) and selects a new robot position based on these evaluations, appending it to the path. If a local minimum is detected, the algorithm resets the path to the starting position and updates observed obstacles. It also checks if the maximum time has been exceeded or if the robot is within the target margin. If any condition is met, the algorithm terminates and returns the path; otherwise, it iterates until finding a satisfactory path or reaching the time limit. \par

The key properties of the different APF algorithms including RAPF are listed in Table~\ref{my-tab:apf_algos}.

\begin{algorithm}
\caption{RAPF algorithm}\label{alg:RAPF}
\small
\begin{algorithmic}[1]
\Require $\mathbf{x}, \mathbf{x}_t, N_{obs}, N_B, StepSize, MaxTime, \{\mathbf{x}_i,r_i\}_{i\in N_{obs}},$
$GoalMargin$
\Ensure $PathToTarget$
\State $\mathcal{N}_{obs}=\{1,2,\hdots, N_{obs}\}$
\State $\mathcal{N}_{B}=\{1,2,\hdots, N_{B}\}$
\State $Path = \mathbf{x}$ 
\State $Timer = CurrentTime$
\State $FindingPath = True$
\While{$FindingPath$}
    \State $J_{a,robot} \gets get\_Ja(Path,\mathbf{x}_t)$
    \State $J_{r,robot} \gets get\_Jr(Path,\{\mathbf{x}_i,r_i\}_{i\in N_{obs}})$
    \State $J_{robot} \gets J_{a,robot} + J_{r,robot}$
    \For{$Idx_{b} \in \mathcal{N}_B$}
        \State $\mathbf{x_b} \gets get\_bacteria(Path,StepSize,N_B,Idx_{b})$
        \State $J_{a,bacteria}\gets get\_Ja(\mathbf{x_b}, \mathbf{x}_t)$
        \State $J_{r,bacteria} \gets get\_Jr(\mathbf{x_b}, \{\mathbf{x}_i,r_i\}_{i\in N_{obs})}$
        \State $J_{bacteria} \gets J_{a,bacteria} + J_{r,bacteria}$
    \EndFor
    \State $new_x \gets pick\_bacteria(Path,J_{robot},J_{bacteria},\mathcal{N}_{obs})$
    \State $Path \gets Add(new_x,Path)$
    \If{$local\_minimum(Path)$}
        \State $Path = \mathbf{x}$
        \State $\mathcal{N}_{obs} \gets Add(\mathbf{x}_{loc\_min},\{\mathbf{x}_i,r_i\}_{i\in N_{obs}})$
    \EndIf
    \If{$Timer - CurrentTime > MaxTime$}
        \State $FindingPath = False$
    \EndIf
    \If{$||\mathbf{x}_t-\mathbf{x}_j||_2 <  GoalMargin$}
        \State \textbf{return} $Path$
    \EndIf
\EndWhile
\State \textbf{return} $None$

\end{algorithmic}
\end{algorithm}




\section{Simulations}

\subsection{Virtual environment}

To test the performance of our algorithms in a more realistic, we perform simulations in a virtual environment consisting of rocks and craters on Lunar surface. We achieve this by studying the size-frequency and cumulative number distributions of these geographical features, which are derived from past lunar mission data. Based on data from \cite{Golombek1997}, the cumulative fractional area covered by rocks with a diameter greater than D versus rock diameter (D) is given by
\begin{equation} \label{eq:rock_cumulative_area}
    F_r(D) = k_re^{-q_r(k_r)*D},
\end{equation}
where $k_r$ is the rock abundance [\%] and $q_r$ [$m^{-1}$] is a coefficient determined empirically. From (\ref{eq:rock_cumulative_area}) the cumulative number of rocks per square meter versus rock diameter ($D_r$) is then given by
\begin{equation} \label{eq:rock_cumulative_number}
    N_r(D) = \frac{4q_r(k_r)k_r}{\pi}\left( \frac{e^{-q_r(k_r)D}}{D} - q_r(k_r)Ei(-q_r(k_r)D)\right). 
\end{equation}
The rock abundance is expected to be smaller than 10\% as observed in previous NASA and ESA missions \cite{Golombek2003, Li2018}. Extending values from previous missions (Apollo, Chang’e, and Luna) to the LZ landing site we obtain $k_r$ = 0.02 and $q_r$ = 1.6 $m^{-1}$. The height-to-diameter ratio allows us to determine that rocks with a diameter greater than $D_{r,crit} = 6.5$cm are considered obstacles for LZ, i.e. a height greater than 35\;mm. The crater abundance model uses the same (\ref{eq:rock_cumulative_area}) and (\ref{eq:rock_cumulative_number}). 
A Monte-Carlo simulation is run to create three different scenarios in which to compare the previously introduced algorithms. The map is 30x30$m^2$ with a distribution of the obstacles within $(x_{min},y_{min}) = (5,5)$ and $(x_{max},y_{max}) = (25,25)$. The rocks cover 1.8\% and the craters 15\% of this 20x20$m^2$ area. The starting position is located at $(x_s,y_s)$ = (2,2) and the target position is a circle centered at $(x_g,y_g)$ = (28,28), with a radius of 0.5m. The radius of the rover is 0.20m. A* is used as the optimal reference for this analysis. Table~\ref{tab:scenario_obstacle} presents the distribution of rocks and craters for each scenario. Despite the large differences between the numbers, the obstacle coverage of the surface is constant as previously defined ($A_{rock}=1.8\%$, $A_{crater}=11\%$). 500 simulations are run on each environment for each algorithm. The position of obstacles is randomly generated for each new simulation.\\

\begin{table}[t]
    \caption{Number of obstacles for each scenario}
    \begin{center}
    \begin{tabular}{|c|c|c|c|}
        \hline 
        \textbf{Scenario} & \textbf{Rocks} & \textbf{Craters} & \textbf{Total obstacles} \\
        \hline
        A & 42 & 38 & 80\\
        \hline
        B & 88 & 32 & 120\\
        \hline
        C & 137 & 24 & 161\\
        \hline
    \end{tabular}
    \label{tab:scenario_obstacle}
    \end{center}
\end{table}

\subsection{Performance metrics}

The following metrics are used to evaluate the performance of the algorithms.

\noindent \textit{Reachability, $R_{s}$:} It defines the ratio of successful navigation trials $N_s$ over $N_m$ attempts. A navigation trial where the rover succeeds in reaching the target without colliding with obstacles defines a successful navigation trial. 
\begin{equation}
    R_{s} = \frac{N_s}{N_m},
\end{equation}
\noindent \textit{Computational complexity, $T_c$ :} It defines the sum of the time the algorithm takes to compute each of the successful paths to the target until the rover reaches it.\\

\noindent \textit{Path length, $M_s$:} It defines the total distance covered by the rover to reach its target. The average path length for $N_{s}$ successful navigation trials is given by
\begin{equation}
    \overline{M_{s}} = \frac{1}{N_{s}} \sum_{k=1}^{N_{s}} M_{s}(k),
\end{equation}
with $M_{s}(k)$ the path length of the \textit{k}th successful run.\\

\noindent \textit{Safety, SF:} Considering the robot has detected \textit{N} obstacles during a successful navigation attempt \textit{k} of length $M_s$ at locations $r_1,r_2,...,r_{M_s}$, the safety parameter refers to the average minimum distance between the robot and the detected obstacles. For $N_s$ successful navigation trials it is defined as
\begin{equation}
    \overline{SF} = \frac{1}{N_s} \sum_{k=1}^{N_s} \frac{\sum_{m=1}^{N(k)}d_{min}(r_o^m)}{N(k)},
\end{equation}
with $d_{min}(r_o^m) = min\{||r_1-r_o^m||,||r_2-r_o^m||,...,||r_{M_s}-r_o^m|| \}$ and $r_o^m$ and the location of the \textit{m}th detected obstacle. Safety will be used to compare field testing with simulation performance of the algorithms.\\

\begin{table*}
\caption{Reachability, mean planning time and mean path length for each scenario and each algorithm}
\label{my-tab:simu_results}
\begin{tabularx}{\textwidth}{@{} l *{10}{C} c @{}}
\toprule
     & \multicolumn{3}{c}{\textbf{Reachability [\%]}} & \multicolumn{3}{c}{\textbf{Mean planning time [ms]}} & \multicolumn{3}{c}{\textbf{Mean path length [m]}}\\
     \textbf{Algorithm} & \textbf{Scenario A} & \textbf{Scenario B} & \textbf{Scenario C} & \textbf{Scenario A} & \textbf{Scenario B} & \textbf{Scenario C} & \textbf{Scenario A} & \textbf{Scenario B} & \textbf{Scenario C} \\
     \midrule
    \textit{A*} & \textit{100} & \textit{100} & \textit{100} & \textit{3906.6} & \textit{5444.3} & \textit{7699.1} & \textit{38.6} & \textit{38.5} & \textit{38.7} \\
    APF & 62.2 & 51.0 & 30.6 & 1225.8 & 1669.6 & 2180.0 & \textbf{39.2} & \textbf{40.0} & \textbf{40.5}\\ 
    RVF & 73.6 & 63.4 & 46.6 & 1634.8 & 2237.5 & 2966.6 & 41 & 41.7 & 42.2\\
    CRBAPF* & 83 & 83.2 & 80 & \textbf{579.6} & \textbf{888.1} & 1328.6 & 43.5 & 44.2 & 45.2\\
    \textbf{RAPF} & \textbf{96.4} & \textbf{93.8} & \textbf{91.8} & 589.6 & 889.5 & \textbf{1276.5} & 39.8 & 40.3 & 41.1\\
\bottomrule
\end{tabularx}
\end{table*}

\subsection{Simulation results}

Simulation results are listed in Table~\ref{my-tab:simu_results}. As anticipated, our reference algorithm, A*, consistently reaches the target in every scenario. The classical APF approach exhibits poor results, reaching the target less than one in three instances as cluttering increases. RVF demonstrates relatively high reachability in non-cluttered environments but experiences a drastic decrease with cluttering. In contrast, CRBAPF* maintains relatively high reachability independent of the environment. Notably, RAPF yields excellent results, exceeding CRBAPF* reachability by more than 15\%.
Planning time for all the algorithms increases with cluttering (from Scenario A to Scenario C), with A* having the longest computation time, as expected. Both RAPF and CRBAPF* share the same planning time, outperforming both RVF and A*. On average, they are 6 times faster than A* and 2.5 times faster than RVF. RVF has a larger computation time than APF as it computes the field of forces and the rotation around each obstacles.\\

\textbf{Stability and outliers}: CRBAPF* have a low variability but a certain amount of outliers, meaning that some configurations may lead to an abnormal computation time. RVF is the most stable and consistent algorithm in terms of planning time as it has a very low variability and almost no outliers. A* presents the same characteristics as RAPF and CRBAPF* in terms of outliers, but shows much higher median and variability, making it less consistent than RAPF. Outliers for CRBAPF* come from the random walk technique implemented to escape the local minima, which our algorithm solves.\\

\textbf{Path length}: In terms of path length, the difference between algorithms becomes less evident. A* and APF yield the shortest paths, but his result may be less reliable for APF, given that the mean path is computed based on successful paths. However, APF exhibits an overall low number of completed paths, indicating that path length is only computed in favorable environments. RVF produces paths between 5\% to 10\% longer than A*. This can be attributed to RVF providing the same direction of rotation to the vector field of clusters of obstacles, leading to detours, while A* finds a direct path between obstacles. CRBAPF* yields a path length between 10\% and 15\% greater than A*, attributable to the random walk technique implemented in CRBAPF*. RAPF generates paths between 3\% and 6\% longer than those computed by A*, significantly lower than the CRBAPF* paths. The path length for A* remains independent of the environment, whereas RVF, RAPF, and CRBAPF* exhibit a slight increase with cluttering. Path length distributions are highly asymmetrical. The median length is lower for CRBAPF* than RVF, while the opposite holds for their means. This discrepancy arises from the substantial and widespread number of outliers generated by CRBAPF*, attributed to the tail effect stemming from the random walk technique. This technique can produce extremely long paths as the path generation relies solely on randomness. \\

\textbf{Planning time}: In terms of planning time, A* presents a large number of outliers but with low scattering. In contrast, RAPF displays half as many outliers and a considerably lower Interquartile Range (IQR) than CRBAPF*, where the random walk technique is not used. The results suggest that RAPF is more consistent between paths, demonstrating increased stability and robustness. The RAPF algorithm displays approximately 50\% lower navigation time, up to 200\% higher success rate, and similar path lengths compared to the conventional APF algorithm. RAPF also outperforms RVF in success rate, path length, and planning time, while the planning time distribution closely mirrors that of the CRBAPF*. Notably, it also outperforms the CRBAPF* in path length, producing a near-optimal path.

\section{Field validation}

\subsection{Testbed description and testing approach}

The rover is tested on a 3x4\;m sandy terrain featuring rocks, craters, and slopes mimicking the lunar surface (see Fig.~\ref{fig:intro}). The bottom left corner defines the origin of the 2D Cartesian reference frame.
A Raspberry Pi 4B 2GB is used as an on-board computer on which the different path planning algorithms are implemented. Six motors drive the six legs. An IMU and an RF Tag are carried on board. The RF Tag communicates with 4 anchors located in each corner of the testbed for localization purposes. A threshold filter is applied to the raw data from the RF Tag to reduce noise. A Kalman filter merges the data from the motor encoders, IMU, and filtered RF Tag data at a frequency of 5Hz. Obstacles are detected with an on-board camera. Objects of a given size are used as obstacles to allow for the estimation of the distance and the angle from the robot. Obstacles are detected in a range of 0.8\;m and a field-of-view of 62°. The software runs on Robotic Operating System (ROS) \textit{Noetic}. Localization data, which consists of the 2D coordinates of the rover ($x_r$, $y_r$) and its heading ($\theta_r$), is shared with the path planning node. The distances, angles, and sizes of all detected obstacles are transmitted upon request from the planning node. 

We completed a set of tests with different starting points, targets, and obstacle layouts. Arrangements that produce local minima are also tested to ensure the robustness of the RAPF algorithm. Tests have been conducted with up to 16 obstacles of varying sizes to comply with the lunar model presented before. Different terrain topologies were also tested (flat surface, overcoming small rocks, and slopes). Twenty tests were performed using virtual obstacle detection. This means that instead of using the on-board camera to detect obstacles, their positions are measured offline and provided to the planning algorithm. The detection is simulated in software as a perfect detection since all potential noise or uncertainty is removed. Simulated object detection assumes the same range and field-of-view of the on-board camera (i.e., 0.8\;m, 62°). This enables us to evaluate the performance of the path planning algorithm irrespective of potential errors induced by other parts of the navigation pipeline. Ten tests were conducted using the on-board camera to prove that RAPF is robust to sensor-induced noise and uncertainties. Details of each test are provided in the following section.

\subsection{Testing results and discussion}

\begin{figure}[htbp]
    \centering
    \includegraphics[width=1.\linewidth]{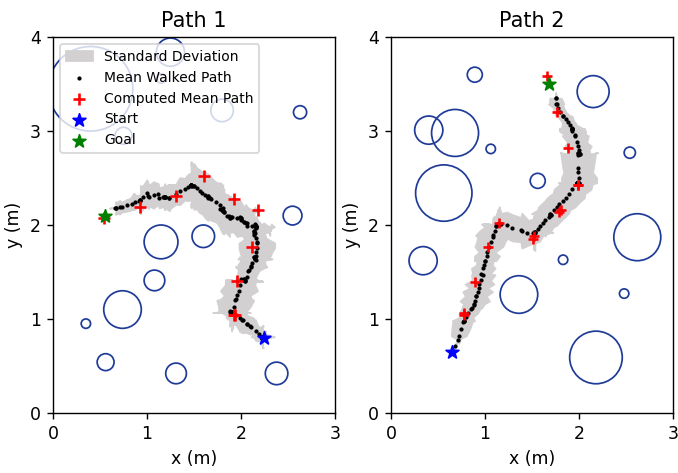}
    \caption{Mean computed path (red) , mean walked path (black) and standard deviation of the mean walked path (gray) for 10 repetitions for two different scenarios (Path1 \& Path 2)}
    \label{fig:ttest_paths}
\end{figure}

\textbf{Robustness}: In a first set of tests, we evaluated the robustness of RAPF against variations in the environment. Ten field tests are performed for two different obstacle layouts. These tests are performed with the virtual obstacle detection configuration only to avoid introducing further uncertainties into the analysis. Obstacles, starting points, and target layouts for the two scenarios, together with the computed and walked paths, are plotted in Fig.~\ref{fig:ttest_paths}.

A t-test is performed to compare the ratio ($\Omega$) of the actual walked path length over the computed path length for two different scenarios, to determine if the RAPF algorithm's behavior is influenced by the environment, with a null hypothesis suggesting no significant difference in the ratio of computed path length to actual walked path length between two map configurations. For a sample size of 10 for each group, a calculated t-statistic of 0.346 and a critical value of 2.110 for a two-tailed test on 17 degrees of freedom fails to reject the null hypothesis, indicating no statistical difference in the analyzed ratio.

The main challenge brought by real-world testing is the uncertainties. In a second set of tests, we evaluated the robustness of RAPF against sensor-induced uncertainties. In this case, data is collected for both rover configurations (virtual and on-board obstacle detection) and across Path 2. A Welch's t-test is employed with a sample size of 10 for each group, resulting in a calculated t-statistic of 0.64 and a degree of freedom of 15. With a p-value of 0.05 and a critical t-value of 2.131, the null hypothesis, suggesting no statistical difference in path length between the two scenarios, cannot be rejected as the calculated t-value is smaller than the critical value.\\
Lastly, it is worth highlighting that the minimum distance maintained between the rover and the nearest obstacle was 7cm on average when the on-board obstacle detection was used and 9cm when the virtual detection was used. The minimum distance kept from all obstacles was 20cm on average for the on-board detection and 26cm for the virtual detection. The hardware detection shows the lowest median and largest IQR, which means that when RAPF is implemented alongside on-board obstacle detection, the robot tends to travel closer to obstacles along its path. The on-board hazard detection leads to a smaller safety distance due to the greater variability in the estimation of the position of the obstacles relative to the rover.\\

\textbf{Computational complexity}: The path planning algorithm is run on the rover on-board computer, a Raspberry Pi 4B 2GB. The total computation time from the starting point to the target for the environments previously defined was always less than 0.2\;s. The time to recompute one path took a maximum of 0.025\;s.

For APF-based path planning methods, memory usage and CPU exploitation are intrinsically related to the size of the potential map. The smaller the map, the lower the overall computational resource needed to compute a path. In APFs, a map of the surroundings of the rover is stored in memory. Every time a new waypoint is reached, all path planning data is cleared from memory and new paths are recomputed. The RAPF algorithm is distinguished by the fact that at bacteria points only the potential is computed. The number of planned paths depends on the level of cluttering of the environment as a new path is recomputed every time an obstacle is detected. For each new planned path, the number of potential values computed can be derived from: 
\begin{equation}
    N_{RAPF} = \frac{N_BM_s}{\rho_B},
\end{equation}
where $N_B$ is the number of bacteria, $M_s$ is the path length, and $\rho_B$ is the distance between the rover and the bacteria point. The RAPF execution speed highly relies, therefore, on the distance between the rover and the targeted waypoint. Our field experiments showed that to reach a waypoint located about 3\;m from the rover, the number of computed potential values for RAPF is 610 on average. In comparison, the estimated number of computed potential values for the grid-based APF and RVF algorithms would be between 1100 and 1300 for a 5x5$m^2$ stored map. Each obstacle is stored in 6 bytes. The size of the list of obstacles stored depends on the number of obstacles $N_o$. From the lunar model derived, up to 35 obstacles can be considered within a map of 5x5 $m^2$. The computed path to follow is also made of \textit{float16} values. It consists of a list of coordinates ($X_r$,$Y_r$), each couple having a size of 4 bytes. The total size depends on the length of the path to follow. For a 5x5 $m^2$ and a waypoint located at around 3\;m from the rover, RAPF approximately requires 0.5\;KB of storage. In comparison, for storing a map of the same size, Field D* needs around 10\;KB of memory \cite{Carsten2007}.\\

\begin{figure}[t]
    \centering
    \includegraphics[width=0.5\textwidth]{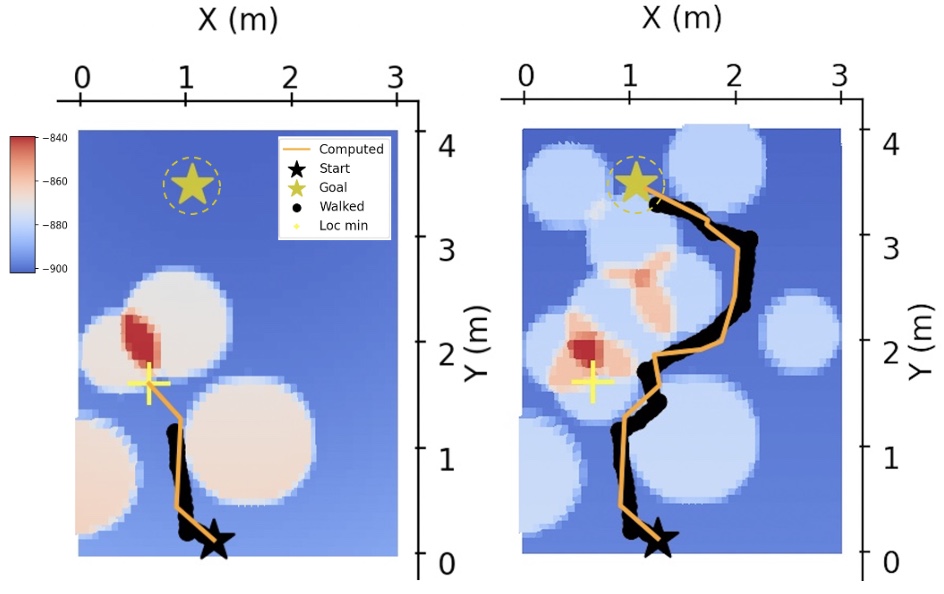}
    \caption{Potential map, computed path (orange), and path followed by the rover (black). Left: CRBAPF. Right: RAPF}
    \label{fig:potential_map}
\end{figure}

\textbf{Reachability}: The path planning module running onboard is capable of estimating the average time it would take to complete a traced path. In combination with the relative location of the rover and the current drawn by each motor, this can help the rover know if it is physically stuck or blocked. The algorithm knows the number of walking and turning steps required to reach a waypoint as well as the number of times a picture needs to be taken to check the path and the time needed to perform these tasks. These data are subject to slippage, actual resource allocation, and terrain uncertainties. Therefore a 100\% margin is added to this estimated time. If the designated waypoint is not reached within the allocated time for executing the path, an error flag is raised and notified to the on-board computer. Data gathered by the rover while navigating are plotted in Fig.\ref{fig:potential_map}. Potential maps as displayed in this figure are not stored on the rover and only computed for bacteria points; only the list of obstacles is stored on board. The potential maps plotted here are for environmental visualization purposes only. Regions between the obstacles and $\rho_0$ normally have infinite potential, represented in these plots by a very large finite value. 
On the left image, no local minima escape method is implemented (i.e., CRBAPF path planner). The rover gets stuck at the first local minimum encountered. On the right image, the RAPF algorithm is used, which implements a potential function assigned to local minima as described before. The path leading to the local minimum is cleared and a new path is computed considering the potential generated by the local minimum. This leads to a new path capable of reaching the target, as can be seen in the figure. In some cases, the algorithm does not manage to plan a path that avoids the local minimum despite the local minima technique implemented. In this case, an error flag '\textit{local\_minimum}' is set by a timer. If no local minimum-free path is output after 10 seconds, the path planning algorithm notifies the OBC that no path can be found. This flag prevents the rover from continuing to search an impossible path and avoids unnecessary resource consumption. 

\section{Summary}

The RAPF algorithm enables real-time computation of new paths for small and resource-limited mobile robots. The proposed approach drastically increases the success rate compared to state-of-the-art APF methods by effectively addressing the local minima issue. The bacterial approach of the RAPF algorithm, minimizes memory requirements, making it ideally suited for robotic systems with strong computational limitations. Simulation results show a 200\% higher success rate and 50\% lower computing time than the classical APF algorithm, with only a 5\% longer path length than the optimal algorithm, A*, in a lunar-like environment. It also outperforms state-of-the-art APF-based algorithms by more than 15\% in reachability and 10\% in path length for a similar or shorter planning time. 
During the field validation, the RAPF algorithm displayed robust performance, effectively adapting to uncertainties introduced by various sensors. The RAPF algorithm performs well in various situations. The rover walks in a straight line towards the target when no obstacles are detected.
It navigates smoothly around small discrete obstacles. The path planning performance of the rover do not depend on the relief of the terrain.  Its consistent performance across diverse environments underscores its capability for real-time autonomous navigation over challenging terrains.

\section*{Acknowledgment}
We express our appreciation to Lunar Zebro \cite{rajan2024} for providing the rover prototype and granting access to their testing facility. The project is partially funded by the Moonshot program supported by the Delft Space Institute.


\vspace{12pt}

\end{document}